\documentclass{sigchi}

\toappear{
Permission to make digital or hard copies of part or all of this work for personal or
classroom use is granted without fee provided that copies are not made or distributed
for profit or commercial advantage and that copies bear this notice and the full citation
on the first page. Copyrights for third-party components of this work must be honored.
For all other uses, contact the owner/author(s).\\\\
© 2023 Copyright held by the owner/author(s). 
}

\usepackage{balance}       
\usepackage{graphics}      
\usepackage[T1]{fontenc}   
\usepackage{txfonts}
\usepackage{mathptmx}
\usepackage[pdflang={en-US},pdftex]{hyperref}
\usepackage{color}
\usepackage{booktabs}
\usepackage{textcomp}

\usepackage{microtype}        
\usepackage{ccicons}          

\usepackage{todonotes}

\usepackage{algorithm}
\usepackage{algpseudocode}

\def\plaintitle{Enhanced Deep Q-Learning for 2D Self-Driving Cars: Implementation and Evaluation on a Custom Track Environment}

\def\emptyauthor{}
\def\plainkeywords{Deep Q-Learning Network (DQN), Self Driving, Neural Network}

\makeatletter
\def\url@leostyle{%
  \@ifundefined{selectfont}{
    \def\UrlFont{\sf}
  }{
    \def\UrlFont{\small\bf\ttfamily}
  }}
\makeatother
\def\pprw{8.5in}
\def\pprh{11in}

\definecolor{linkColor}{RGB}{6,125,233}
\hypersetup{%
  pdftitle={\plaintitle},
  pdfauthor={\emptyauthor},
  pdfkeywords={\plainkeywords},
  pdfdisplaydoctitle=true, 
  bookmarksnumbered,
  pdfstartview={FitH},
  colorlinks,
  citecolor=black,
  filecolor=black,
  linkcolor=black,
  urlcolor=linkColor,
  breaklinks=true,
  hypertexnames=false
}

\begin{document}

\title{\plaintitle}

\numberofauthors{3}
\author{%
\alignauthor{Sagar Pathak\\
    \affaddr{Computer Science}\\
    \affaddr{University of Memphis}\\
    \email{spathak1@memphis.edu}}\\
  \alignauthor{Bidhya Shrestha\\
    \affaddr{Computer Science}\\
    \affaddr{University of Memphis}\\
    \email{bshrstha@memphis.edu}}\\
  \alignauthor{Kritish Pahi\\
    \affaddr{Computer Science}\\
    \affaddr{University of Memphis}\\
    \email{kpahi@memphis.edu}}
}

\maketitle

\begin{abstract}
  This research project presents the implementation of a Deep Q-Learning Network (DQN) for a self-driving car on a 2-dimensional (2D) track, aiming to enhance the performance of the DQN network. It covers the development of custom driving environment with pygame on the track around the University of Memphis map and design and implementation of the DQN model. The algorithm utilizes data from 7 sensors collected by sensors installed in the car, based on the distance between the car and the track. These sensors are positioned in front of the vehicle, spaced 20 degrees apart, enabling them to sense a wide area ahead. We successfully implemented DQN and also modified version of DQN with priority-based action selection mechanism and referred it as modified DQN. The model is trained on 1000 episodes and the average reward received by agent is found to be around 40 which is around 60\% higher than the original DQN and around 50\% higher than the vanilla neural network.

\end{abstract}

\ccsdesc[500]{Human-centered computing~Human computer interaction (HCI)}
\ccsdesc[300]{Human-centered computing~Haptic devices}
\ccsdesc[100]{Human-centered computing~User studies}

\keywords{\plainkeywords}


\section{Introduction}
Self-driving, also known as autonomous driving, aims to navigate from a source point to a destination point without human intervention. Tasks involved in self-driving include lane tracking, maintaining speed, navigating obstacles, keeping a safe distance from other vehicles, and progressing towards the destination. In essence, a self-driving car must operate effectively in dynamic and unpredictable environments where road conditions and traffic patterns can frequently change. With the increasing adoption of various machine learning frameworks such as Neural Networks, artificial intelligence (AI) agents are capable of controlling and driving vehicles autonomously. Self-driving cars represent one of the most promising application areas of reinforcement learning.

\begin{figure}[!h]
\centering
  \includegraphics[width=0.9\columnwidth]{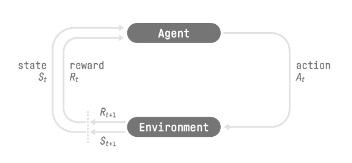}
  \caption{Flowchart of reinforcement learning}~\label{fig:figure1}
\end{figure}

\subsection{Reinforcement Learning}
Reinforcement Learning is a machine learning technique that enables an agent to learn by interacting with the environment. The agent makes decisions to take actions based on rewards and penalties it gets from taking a particular action in a particular states. It is an interactive process by trial and error using feedback from its own actions and experiences. The goal is to find a suitable action model that would increase the total cumulative reward of the agent. The agent evaluates a current situation state (St), takes an action (At), and receives rewards (Rt) from the environment after each action At as shown in Fig 1. Reward can be positive feedback as well as negative feedback or punishment for making a mistake.

Q-learning is one of the popular reinforcement learning algorithms used to solve a given problem in which an agent makes a decision based on the reward received for each action taken. The algorithm updates the Q values which denote the value of performing action a in state s. The goal is to learn an action-value function also known as a Q-function which estimates the maximum future reward for taking action given a state s. The calculation of the Q-value function takes into account the maximum discounted future reward for an agent to move from state s to state s’. The Q-value function can be iteratively converged to the optimal Q-value function by the difference of the estimated utilities between current state st, and next state st+1, with learning rate alpha, at time step unit t, as shown in equation \eqref{eqn:bellman1} and \eqref{eqn:bellman2}.
\begin{quote}
    \begin{equation}\label{eqn:bellman1}
        A = r_{t+1} + \gamma \cdot \max_{a} Q_t(s_{t+1}, a)
    \end{equation}
    \begin{equation}\label{eqn:bellman2}
        Q_{t+1}(s_t, a_t) = Q_{t}(s_t, a_t) + \alpha(A - Q_t(s_t, a_t)) 
    \end{equation}
\end{quote}

With big state spaces, Q-learning remains ineffective because it would require equally large storage space to store Q-values for each action (a) in state (s). Also, Q-learning cannot be used in unknown states because it can’t infer the Q-values of new states from the previous ones. As we have previously mentioned, self-driving in a dynamic environment comprises a lot of states that are not clear beforehand. Therefore, we cannot rely on Q-learning to train the agent.
\subsection{Deep Q-Learning Network (DQN)}
DQN overcomes these limitations in Q-learning. The algorithm uses a neural network to approximate the Q value function. The current state from the environment is passed as input to the neural network and outputs the Q values for all possible actions and we take the biggest as our next action. 
\begin{quote}
    \begin{equation}
    \label{eqn:dqn}
        Q(s_t, a_t) = R_{t} + \gamma \cdot \max_{a} Q(s_{t+1}, a) 
    \end{equation}
\end{quote}
In a DQN, the agent holds a memory buffer with all past experiences. 
The next action the agent takes is determined by the maximum output (Q-value) of the network. The loss function used in DQNs is the mean squared error of the predicted Q value and the target Q-value. The target Q-value is calculated using the Bellman equation, which states that the target is the immediate reward (Rt) plus the discounted value (g) of the maximum Q-value of the next state as shown in equation \eqref{eqn:dqn}.

\subsection{PyGame Learning Environment}
PyGame Learning Environment is an open-source Python Library that provides a platform for developing and testing reinforcement learning (RL). It is a learning environment that simplifies the gaming environment. It mimics the arcade game environment by providing an interface where the agent can act on the environment with actions. The goal of PLE is to allow practitioners to focus on the design of models and experiments instead of environment design.

\subsection{Motivation}
Training an AI agent that drives autonomously in a real-world scenario is expensive in terms of time and equally dangerous as it involves crashes. Also to train the agent with supervised learning, we need tons of real-world data which might not be feasible in autonomous driving in a dynamic environment. There it is essential to have a simulator that we can run to train the agent that can self-drive in any given track. In this project, we designed a simple game environment as a simulator to train our model to drive along the path using sensor data as input. We focused on understanding the fundamental principles of autonomous driving in a dynamic environment rather than using traditional rule-based systems or pre-programmed algorithms.

Throughout the project all of the team members equally contributed to the work. Each team member were assigned to lead and collaborate in different areas of the development work such as Game environment development work, DQN implementation, Prioritization algorithm, Experimentation, Evaluation, and Documentation.

\section{Related Work}
The state-of-the-art work done by Andy Wu (2020) \cite{research1} on car racing using OpenAI Gym toolkit implemented Deep Q Learning Network (DQN) for the learning environment with TensorFlow and Keras library. The Q value for each action is outputted by the Deep Q Network (DQN) when 3 consecutive top views of the current state of the 2d car racing game are taken as input. The input shape was of 96x96x3, where 3 means the consecutive 3 latest views of the current state. This work used three convolution layers for extracting features from the image and two max pooling layers for preserving important features while discarding unnecessary information. The action space consists of 12 discrete actions where three states for steering left, right and straight, two states for the break with 20\% break or release break and two states for the gas with full gas or release gas actions. In this learning process, the latest 3 states of 96x96 pixels grayscale images are stacked by the program and fed to the trained model, which produces the Q values for the 12 actions to make the car self-driving during the game. The action with the highest Q value is chosen by the agent, which would obtain the most rewards in response to the environment if the model is well-trained and then these steps are repeated until some stoppage criteria. The author tested the agent's behavior in different episodes. The analysis is done in three different training episodes which were 400, 500 and 600. After 400 episodes of training, it is found that the learning agent knew how to take the shortcuts on the map but it was failing to take the sharp right turn and stuck on it. In 500 episodes of the training, the learning agent performed better as compared to 400 episodes and it was driving faster and smoother with few mistakes. But after 600 episodes, the agent become greedier and learned to drive recklessly and mostly going off the track when reaching sharp turns. The average reward collected after the 350 episodes was 550 and the score reached to 200 in 600 episodes. The epsilon was decaying faster and reached to the minimum epsilon at 140 episodes. The author found challenges on balancing the state of the agent in sharp turns and narrow tracks and considered as the future works. 
\newline
\newline
\noindent
Another research by Yang Thee Quek, Li Ling Koh et. al. (2021) \cite{research2} in simulated autonomous vehicle control using deep Q-networks presented the implementation of DQN in two different simulated environments for the self-driving vehicle control.  One of the environments was a simple 2D environment and the other was a 3D environment using Unity software. 
The virtual car agent learns and adapts to its surroundings based on the scores and pixel inputs, and devises the optimal solution for moving the car from point to point on a virtual highway. The study concludes that DQN is a useful learning strategy for the agent of an autonomous vehicle. Without prior knowledge of the surroundings, the autonomous vehicle eventually learned maneuver operations and gained the ability to properly traverse and avoid obstacles in 2D as well as 3D simulated scenarios. They also used Double DQN to mitigate the issue of negative effects on the quality of the resulting policy. Double DQN helped to reduce the effect of Q values overestimation by diving the operations in the target into evaluation selection and action selection. They created three stages for the learning agent named as observe, train, and test. The observe basically observes the environment and takes the state, action reward pair which will be stored in the replay memory. The speed of the car is determined by diving the current speed with the maximum attainable speed in the network which calculates the reward function. The reward function is created in such a way that if the agent gets collide with the path or other object, it will be penalized by -10 and if the speed is zero then it will be penalized by -1 and otherwise the reward will be calculated using the minimum of 10 or 0.2+5*S, where S is the speed at the current step. In the training stage, the actions will be selected using the epsilon greedy approach and the convolution neural network will predict the next action for the particular state and loss will be calculated by squaring the difference between the target value and the predicted value. In the test phase, the agent will not explore but exploits 100 percent of the environment. The image captured from the environment was of 512x288 pixels which was RBG image and it was very hard to work on such high pixels images thus they applied the image dimension reduction to 80x80 and also converted the RBG image to a grayscale image, because the color is not important for the learning agent in car racing environment in that project. After reducing the image dimension, they normalized the pixel values from (0,255) to (0,1) so that the computation will be easier and faster. That input is fed into the convolution layer where they applied the stride of 4. They have a few convolution layers along with the max pooling layers and also the fully connected layers at the end. The final fully connected layer produces the actions that the agent takes. They trained the agent for various rounds and it is shown that in the initial stage of the DQN learning, the agent acted randomly and kept colliding with the obstacles but later on as the Q-values increase along with each timestep, the agent was performing well and stabilized. The accumulated reward trend was in upward trending and in 487 episodes it was seen that the total reward gained by the agent was more than 25. In the completion of the training, the agent achieved a total reward up to 60. They also showed the statistics of the Q values accumulation based on the time and it is seen that they were able to collect the Q values in different time frames such as 2h, 4h, 6h and 8h, where h represents the hour and Q values has slight changes after 4 hours of training which means the reward and actions affect the Q-values as previously theorized. In conclusion, the agent they trained obtained effective control methods and effectively controlled the vehicle in both situations, as demonstrated by the results of the experiment provided in their study. After hours of practice, the agent learned to handle the vehicle such that it could retain its full speed while avoiding obstructions. As the future work they wanted to try different sensors integrated into the simulation of the  2D and 3D vehicle environment using the Double DQN algorithm. 
\newline
\newline
\noindent
The research conducted by Tai Lei and Liu Ming (2016) \cite{research3} on a robot exploration strategy using a Q-learning network implemented the corridor environment for the robot with the depth information from the RGB-D sensor. This research is useful for our car racing environment as well because the car agent also needs to explore the track in an optimized and better way. They also used the convolutional neural network approach where there are three convolutional layers, three maxpool layers, and softmax and ReLu activation functions were used. The input images will be downsampled and fed into the model. The feature maps of every input will be fully connected and fed into the last softmax layer which produces the output. They used the Gazebo simulated 3D environment where the robot will explore, the Q-learning is used for the learning framework and the Turtlebot is used as an agent. The training parameters were the batch size, replay memory size, discount factor, learning rate, gradient moment, max iteration, step size and gamma. They have two different corridors where one is straight and the other is the circular. The reward function is designed in such a way that if it collides or stops it will be penalized by -50 and if it keeps moving then in each step it will be rewarded as 1 which will be accumulated for the episode. They trained the agent for 15000 iterations in straight as well as the circular corridor. It is seen that the loss is decreasing very sharply till 1000 iterations and then stabilized around in the range of 20 for the straight corridor setting whereas the training loss is decreasing for circular corridor event after the 12000 iterations. This shows that the agent is still learning even after the 15000 iterations of training. In conclusion, they confirmed that the Q-learning will be very reliable to achieve obstacle avoidance ability in the simulated environment. They wanted to experiment with DQN framework in real-time exploration directly using GPUs and also wanted to design the new reward function as the future work. 
\newline
\newline
\noindent
Ahmad El Sallab, Momammed Abdou, Etienne Perot and et. al. (2017) \cite{research4} proposed a deep reinforcement learning framework for autonomous driving. The autonomous driving agent was developed considering three different aspects named as: recognition, prediction, and planning. The recognition is a process of identifying surrounding components in the environment which can be pedestrian detection, traffic signs etc. The prediction is the process where the framework builds internal models and predicts the future states and the planning phase was for the integration of recognition and prediction tasks. 

\begin{figure}[!h]
\centering
  \includegraphics[width=0.9\columnwidth]{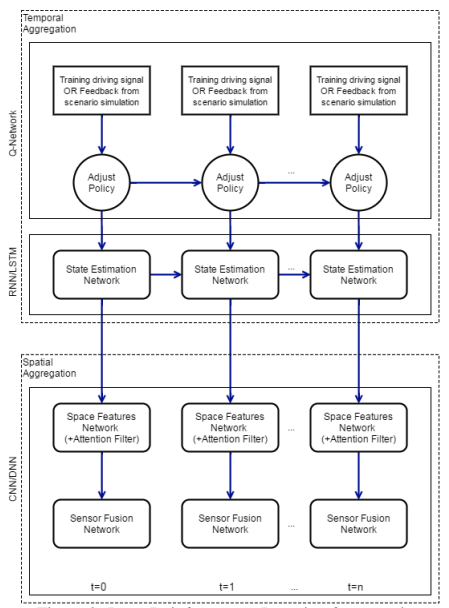}
  \caption{Proposed Deep RL Framework for autonomous driving by Ahmad El Sallab et. al.}~\label{fig:figure2}
\end{figure}

The deep reinforcement learning framework they proposed has two levels of aggregation, one is the spatial aggregation and the other is temporal aggregation. In the spatial aggregation, they created two networks one is for the sensor fusion and the other for the spatial features. The sensor fusion network has the Deep Neural Network (DNN) which is responsible for extracting relevant features. Grouping all sensor information into a raw input vector for the DNN to fuse through weight adjustment using Stochastic Gradient Descent (SGD) is effective, with gradients used to weight each sensor feature based on its relevance to minimizing the overall cost function. A CNN is best suited for this part of the process. They used Long Short Term Memory (LSTM) networks for spatial features processing. In the recurrent temporal aggregation block, they have state estimation network where the driving signals were trained or the feedback from the scenario is simulated and analyzed to adjust the policy. In the final part of the pipeline (i.e planning), reinforcement learning is performed using Deep Q-Network (DQN). They used lane keeping assist algorithm on Torcs which is a open source racing car simulator. The sensor gives the trackPos which fed into the network and gives the track border, speed and x-position. For the optimization process, they used Deep Deterministic Actor Critic (DDAC) network which is most suitable for continuous state cases which helped to smooth the actions and provided better performance. Also, they removed the replay memory trick that helped faster convergence and better performance. This work gave a summary of recent achievements in Deep Reinforcement Learning, followed by a proposal for an end-to-end Deep Reinforcement learning pipeline for autonomous driving that included RNNs to account for Partially Observable Markov Decision Process (POMDP) situations. Attention models were included into the framework, directing CNN kernels to parts of the input crucial to the driving process, reducing computing complexity. The system was evaluated for the lane keep assist algorithm. Future study will include installing the proposed framework in a simulated environment, perhaps extending it to real-world driving scenarios.
\newline
\newline
\noindent
The Deep Q-Network based decision-making for autonomous driving by Max Peter Ronecker and Yuan Zhu (2019) \cite{research5} proposed methodologies for safely navigating an autonomous vehicle in highway scenarios. They proposed a system for lane change maneuvers where the decision-making and timing will be drawn by the deep reinforcement learning policy.  
\begin{figure}[!h]
\centering
  \includegraphics[width=0.9\columnwidth]{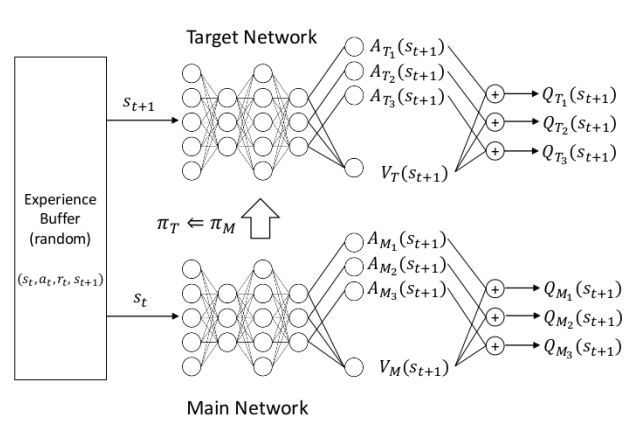}
  \caption{Double DQN structure proposed by Max Peter et. al. }~\label{fig:figure2}
\end{figure}
They implemented two neural networks, one target network and a main network to form a combined double DQN where the experience buffer will be processed. One of the networks in the double DQN will predict the action and the other network will evaluate the action. The developed technique for assessing a Deep-Q-Network backed by conventional control theory was applied to two distinct situations in which the training process, as well as quantitative and qualitative results, were assessed. The system was tested with three different inputs, including the state, a limited view, and a grid map, with the overtake scenario requiring the agent to take over multiple slower vehicles and return to the initial lane, and the second scenario requiring the agent to avoid blocking or crashing into other vehicles, with policies evaluated over multiple random seeds to provide a statistically relevant result due to a common issue in deep reinforcement learning. Future research topics include the design of more complex state and action spaces to overcome limitations, reward design, combination with online planning, and transitioning from simulation to the environment to improve autonomous vehicle capabilities and results. The proposed system produced efficient and safe driving behavior.  

\section{Reinforcement Learning (RL) Environment}
The RL environment for self-driving is developed in Pygame \cite{pygame}. The essential components in our environment are the car, map/track and sensors of the car. The car in our gaming environment is loaded from an image as a Sprite object. Sprite represents a 2D image or animation that can be displayed on the screen. Similarly, the map is also loaded as a Sprite object from an image. We manually drew the map resembling the University of Memphis location. When we load the map image, the transparent segment of the image is masked as the track and the opaque portion as the obstacle. The main reason for using sprite objects in the game is to handle collision detection. Pygame provides inbuilt methods to detect if there is collision between the sprite objects or sprite groups. This has made it easy for us to detect collisions of the car with the track during the game. Similarly, the sensor of a car is drawn as a sprite object using pygame built-in draw line function. The 7 sensors are placed in front of the car at 5 degrees apart from each other. The sensors are placed at the front because our car moves forward only.

In order to simplify our gaming, we made our car move forward with constant speed. In order words, we removed both throttle and braking controls for the car. Our car only has two controls: Steer Left or Steer Right. At the beginning of development, we added both acceleration and braking controls but when we introduced DQN to drive the car, we needed to maintain additional action space for acceleration and braking. This made our experimentation more random so we reduced these controls from the car.

The core of the game is to update the car’s position with the user's control and the collision of the car with the track. As we mentioned earlier, the users are provided with two controls: to steer left and to steer right. The car’s position is updated as it has constant speed and direction is changed with these controls. While updating the car’s position, we constantly check if it collided with the track. The collision is based on the sprite of the map overlapping with the sprite of the car. Pygame provides a built-in function to check this overlapping which returns a boolean flag. If the flag is true, we conclude that the car has collided with the obstacle and thus end the game.

The sensors of the car work in a similar fashion. Each sensor represents a distance value from the car to the obstacles in the track in their respective direction. To compute distance, we first created a line surface using pygame inbuilt function to draw lines. The length of distance is increased by a factor of 1 from the car position until there is collision with the obstacle in the map. The length of the line can maximum be 1000 as we used a loop to increase the line segment of the sensor. Finally, when returning those lengths as the state for the reinforcement learning environment, we normalized the length value by dividing the current values by 1000. This helped us limit the observation state while implementing DQN.

\section{SUMO Framework for Simulation}
The Simulation of Urban Mobility (SUMO) \cite{sumo} is a traffic simulation framework that works at the microscopic level and is open-source. It allows the modeling and analysis of traffic scenarios, including individual vehicle movements, interactions with the road network, and the impact of traffic management policies and strategies. 
\begin{figure}[!h]
\centering
  \includegraphics[width=0.9\columnwidth]{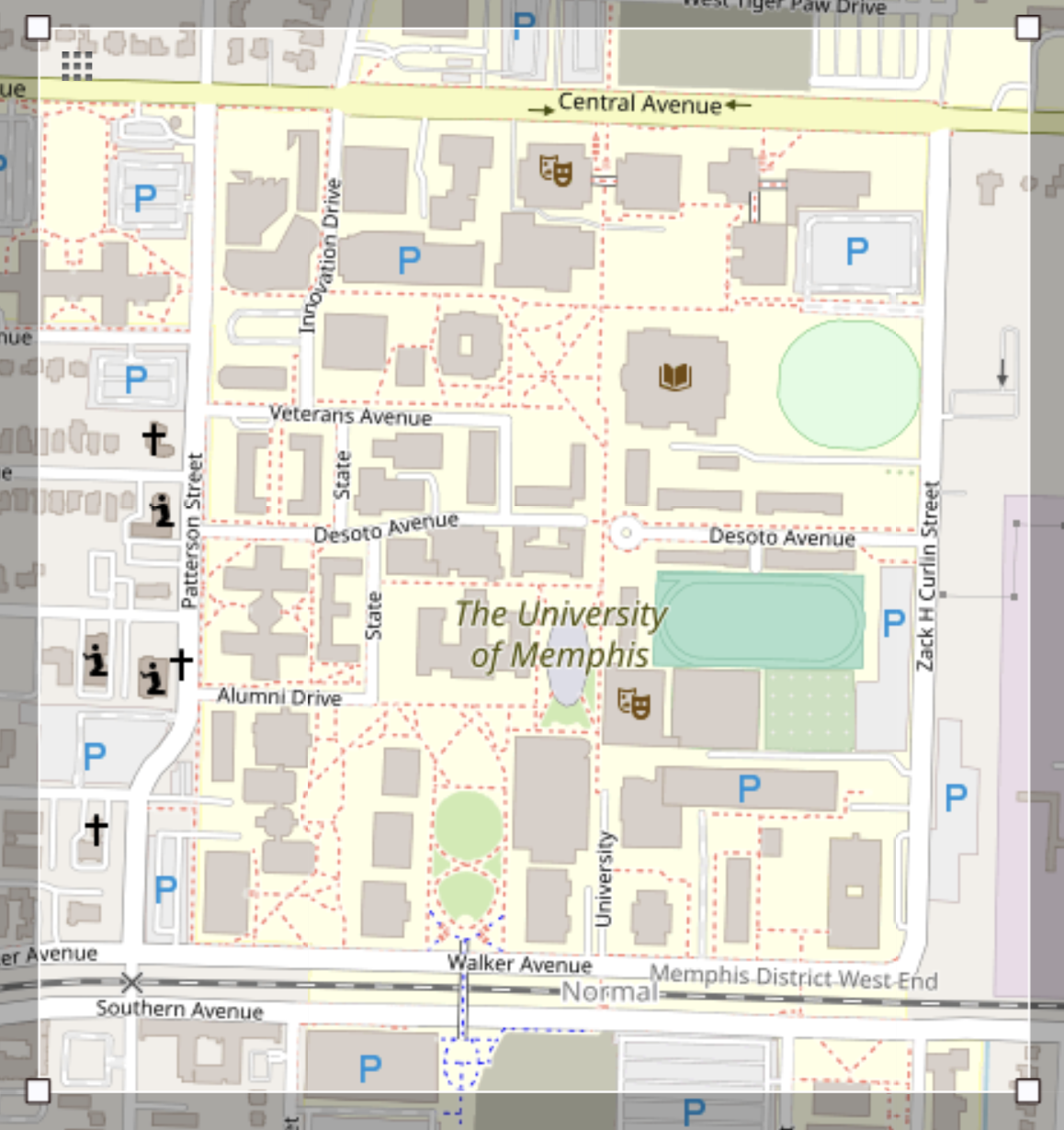}
  \caption{Selected region for the vehicle simulation around the University of Memphis in OpenStreetMap}~\label{fig:game}
\end{figure}
SUMO provides various tools and modules to simulate different aspects of urban mobility, such as vehicle routing, public transportation, pedestrian behavior, and emissions. It can be used for research, education, and transportation system planning and can be easily customized to fit specific needs. Additionally, SUMO can be integrated with other simulation frameworks and tools. We intended to use the SUMO framework to simulate traffic with multiple vehicles around the University of Memphis map. However, due to time constraints, we could only partially utilize the framework due to its complexity. One useful feature of the framework is the ability to extract a real map from OpenStreetMap \cite{osm} and convert it into an XML-based format using the netconvert \cite{netconvert} tool. We then processed the converted XML map with the duarouter \cite{duarouter} library to create routes within the map. To create random trips within the captured map, we utilized the randomtrips \cite{randomtrips} library provided by the SUMO framework. 

\begin{figure}[!h]
\centering
  \includegraphics[width=0.9\columnwidth]{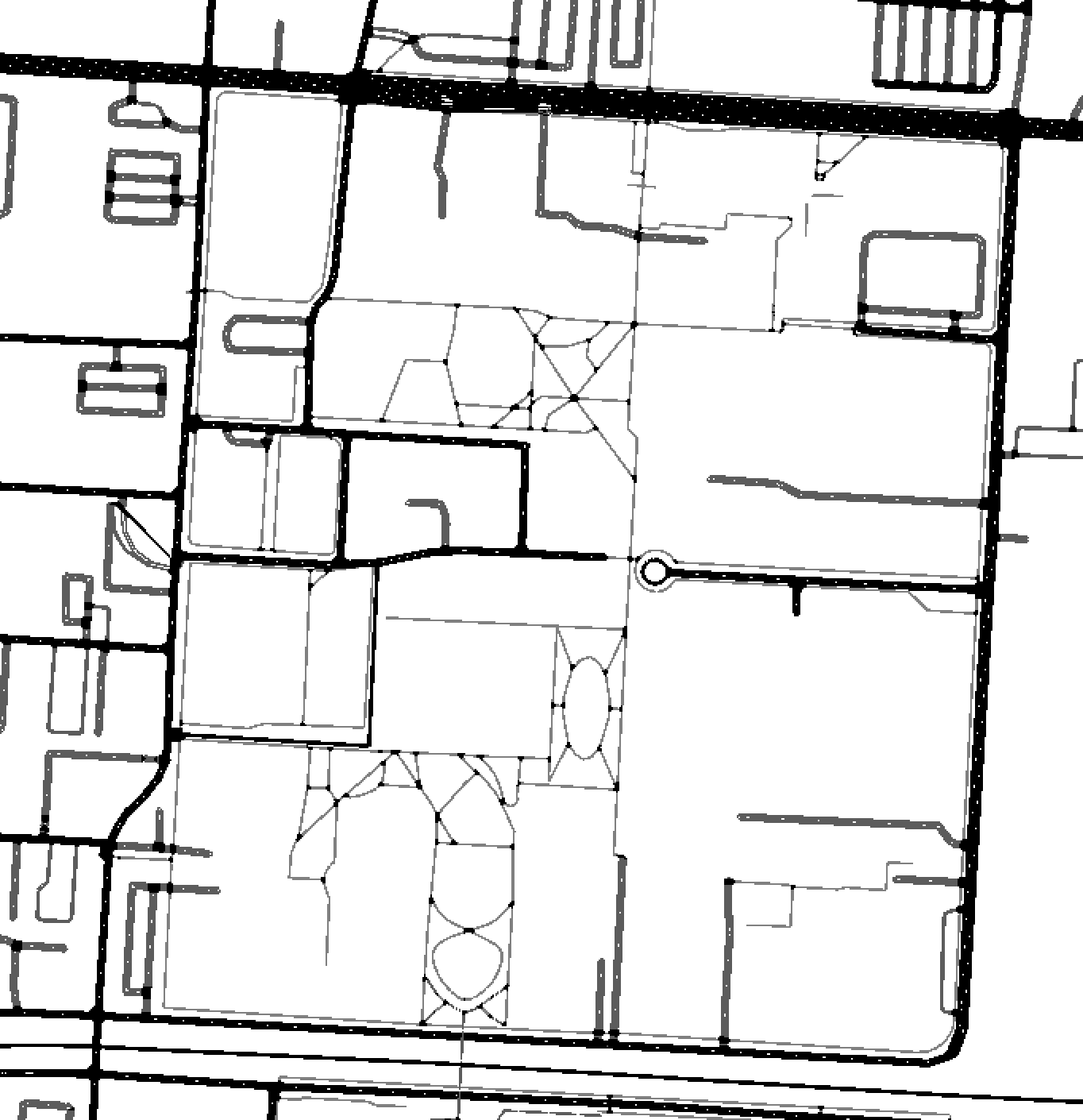}
  \caption{Network Map after converted using SUMO netconvert tool}~\label{fig:game}
\end{figure}

Finally, we linked the trips and routes in the SUMO config \cite{sumoconfig}, and performed the simulation using the SUMO GUI \cite{sumogui} tool. It should be noted that while the SUMO framework can be challenging to work with, it provides many powerful tools and libraries for simulating various aspects of urban mobility, making it a valuable resource for transportation planning and research. 

\begin{figure}[!h]
\centering
  \includegraphics[width=0.9\columnwidth]{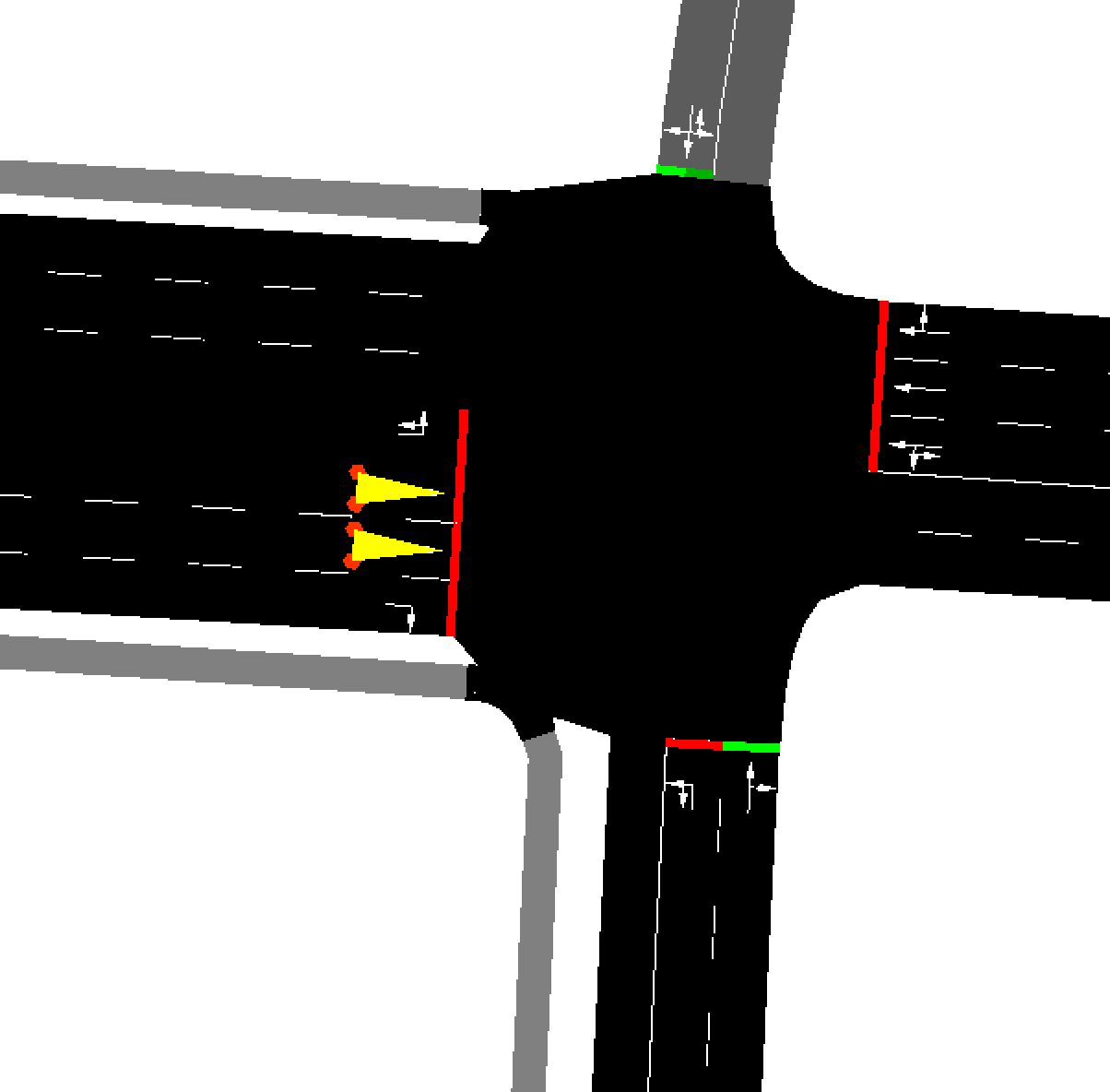}
  \caption{Screenshot of the simulation in Central Avenue and Zack Curlin Street junction near University of Memphis}~\label{fig:game}
\end{figure}

Due to the limited time for this research work, the TraCI \cite{traci}, traffic control interface took time learn and implement into our reinforcement environment. As a result, we were only able to complete the simulation using the SUMO framework on the University of Memphis map. We plan to explore the integration of TraCI in future work. However, the map that we used for our learning purpose was created using Photoshop which mimics the actual map.

\section{Methodology}
In our environment, we have 3 discrete action spaces. As we have set a constant speed for the car, we only needed to steer left or right. However, we added another action which is to do nothing and allowing the car to constantly move forward. Another reason for adding this action is to allow the DQN agent to drive the car more precisely with steering much to the left and right direction just to move forward. If this action is not added in our environment, the model would have to continuously turn in any particular direction, even when moving forward.

\begin{figure}[!h]
\centering
  \includegraphics[width=0.9\columnwidth]{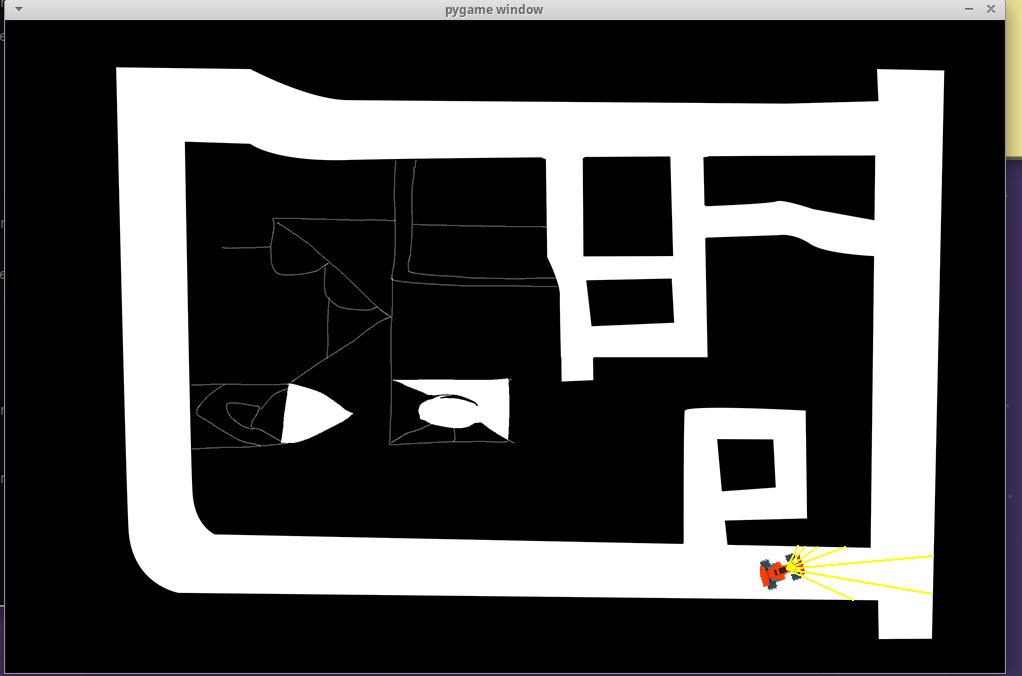}
  \caption{Self driving car in the environment with sensors. }~\label{fig:game}
\end{figure}

The observation space of our environment is the 7 values from the sensors in the car. These are positive normalized distance values of the space between the car and the obstacle in the track. The getGameState method returns these normalized values. These values are changed every step when an action is taken i.e when the position of the car is updated as shown in figure \ref{fig:game}. Our reward function is a simple score which adds +5 if the car doesn’t crash with that action otherwise -20 if the car crashes. The cumulative reward is the overall score during the game as shown in algorithm \ref{alg:rew}.

\begin{algorithm}
\caption{Reward function }\label{alg:rew}
\begin{algorithmic}
\For{ each action taken}
\If{no collision of car with obstacles in map} 
    \State reward = +5 
\Else
    \State reward = -20
\EndIf 
\State score += reward
\EndFor

\end{algorithmic}
\end{algorithm}

\subsection{DQN Implementation}
Deep Q-Network (DQN) algorithm was implemented using the TensorFlow Sequential model. The algorithm extends Q-learning by using a deep neural network to approximate the Q-function, which estimates the expected future rewards for each possible action at each state. The DQN model consists of three main components: a replay buffer, a target network, and a main network. The replay buffer is a memory that stores experiences (i.e., state-action-reward-next-state tuples) during the agent's interactions with the environment. The target network is a copy of the main network that is periodically updated to reduce the correlation between the target and main values. The main network is the neural network that approximates the Q-function and is updated during the training process. The DQN algorithm follows an epsilon-greedy exploration strategy, where the agent chooses the action with the highest Q-value with a probability of epsilon, and a random action with a probability of epsilon. The epsilon value is gradually decreased over time to encourage the agent to exploit its learned policy. The training process for the DQN model involves iteratively sampling experiences from the replay buffer, updating the main network using a loss function that minimizes the difference between the predicted and target Q-values, and periodically updating the target network. The algorithm implementation for DQN is shown in \ref{alg:dqn}.

\begin{algorithm}
\caption{DQN algorithm }\label{alg:dqn}
\begin{algorithmic}
\Require replay buffer of size N
\Require set epsilon and decay rate
\Ensure Initialize the main network
\Ensure Copy the main network to create a target network
\For{each episodes $\leq 1000 $}
    \State reset game state
    \State take do-nothing action and get game state
    \While{ not game\_over}
        \State choose action from the current game state
        \State update game state
        \State collect new\_state, reward, game\_over flag
        \State add the experience to the replay buffer
        \State sample a batch of experience from the replay buffer
        \State train the main network using the minibatch
        \State update the target network using main network 
        \State decay the epsilon value
    \EndWhile
    \State save the model
\EndFor
\end{algorithmic}
\end{algorithm}

The DQN model we implemented has a 3 dense layer and input dimensions of 7 and output dimension of 3 as shown in Table \ref{dqn-arch}. Also, our hyper parameters are shown in Table \ref{hyp}.
\begin{table}[!h]
\begin{center}
\begin{tabular}{ |c|c|c| } 
\hline
Layer & Shape & Parameters \\
\hline
dense & 64 & 512 \\
dense\_1 & 64 & 4160 \\
dense\_2 & 3 & 195 \\
\hline
\end{tabular}
\caption{\label{dqn-arch} DQN layers}
\end{center}
\end{table}

\begin{table}
\begin{center}
\begin{tabular}{ l r  } 
\hline
Epsilon: & 0.99 \\
Discount Factor: & 0.97 \\
Memory Size: & 3000 \\
Replay Memory Size: & 500 \\
Mini Batch Size: & 128 \\
Min Epsilon: & 0.001 \\
\hline
\end{tabular}
\caption{\label{hyp} Hyper Parameters}
\end{center}
\end{table}

\subsubsection{Priority Algorithm}
In our implementation, we further tweak the action selection method based on the model prediction. During the exploitation strategy, we add priority factor to the model output value. The algorithm for boosting the priority factor is shown in \ref{alg:prio}.

\begin{algorithm}[!h]
\caption{Priority based action selection algorithm }\label{alg:prio}
\begin{algorithmic}
\Ensure Feed sensors data to model and get output
\If{left sensor data is greater} 
    \State prioritize steering to LEFT
\ElsIf{right sensor data is greater} 
    \State prioritize steering to RIGHT
\Else
    \State Do nothing/Go straight
\EndIf
\end{algorithmic}
\end{algorithm}

\section{Experiments and Results}
The experiment was conducted on Lenovo Thinkpad E14, (16GB RAM) with Ryzen 5-4500U and Macbook Pro M1 (16GB RAM) with 16-core GPU. We analyzed the performance of DQN, DQN with priority based action selection algorithm and neural network for self-driving car using rewards received per episode. 
\newline

We started the experiment with DQN and CPU, but the car could not learn well. It took a very long time even to travel a short distance. Then we experimented with vanilla neural networks. It was observed that the vanilla neural network performed quite well. However, our ultimate goal was to teach the car to move from source to destination using DQN so we optimized it by introducing priority based action selection algorithm. This method performed well. And with GPU the training time was also much faster. It took around 4 hours to complete 1000 episodes while the training time using CPU was around 12 hours.

\begin{figure}[!h]
\centering
  \includegraphics[width=0.9\columnwidth]{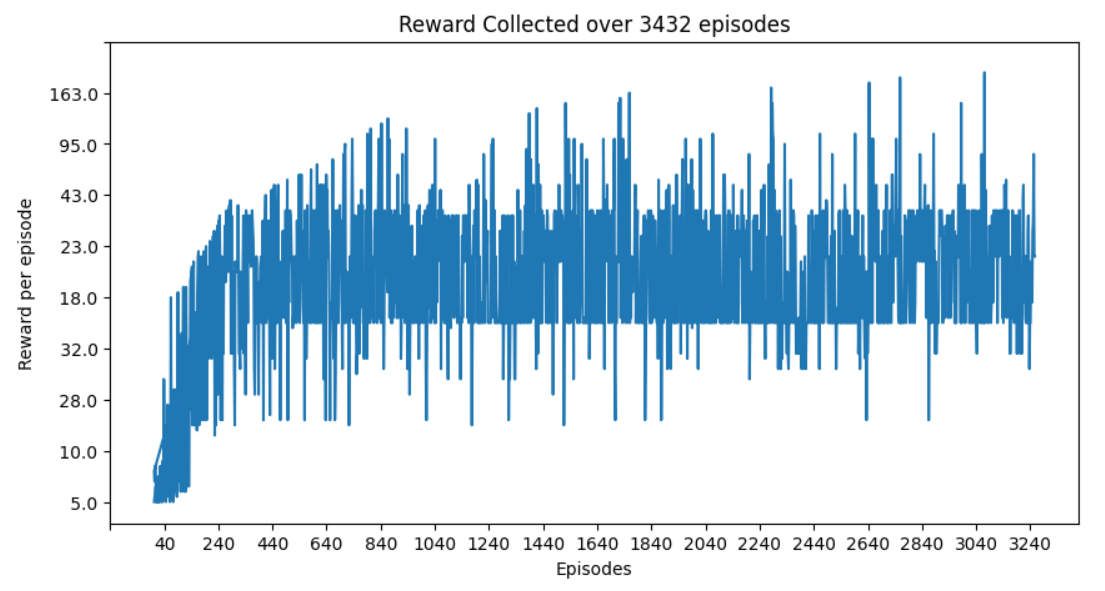}
  \caption{Performance of vanilla neural network}~\label{fig:figure1}
\end{figure}
\begin{figure}[!h]
\centering
  \includegraphics[width=0.9\columnwidth]{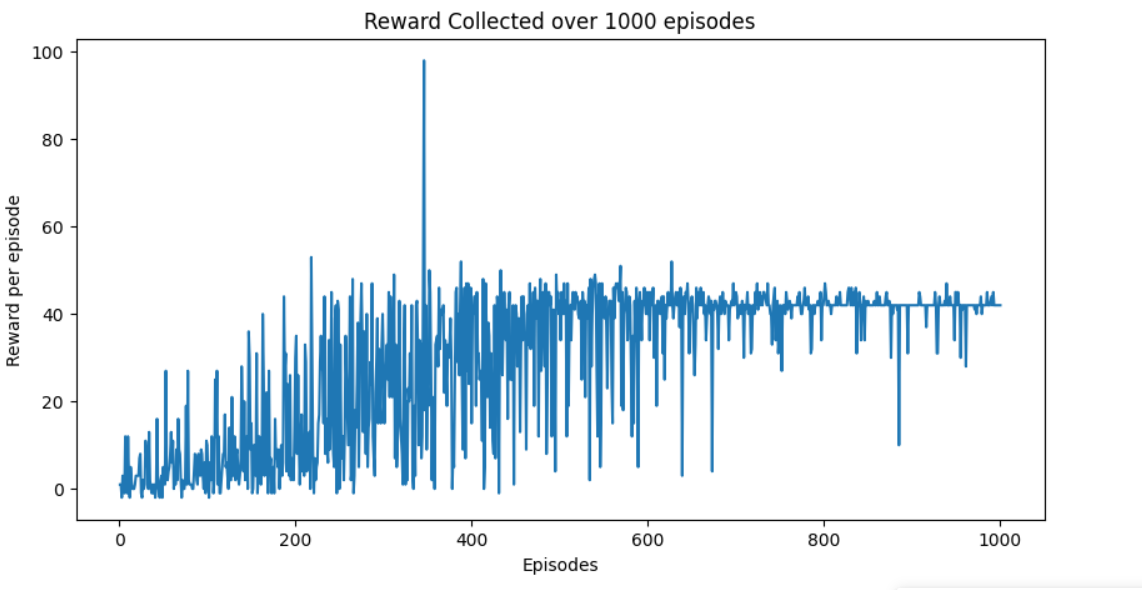}
  \caption{Performance of modified DQN }~\label{fig:figure1}
\end{figure}

\section{Discussion}
The experiment showed it was hard for the original DQN to finish reach from source to destination which is to complete the whole track. However tweaking the action selection mechanism improved the performance and the car could make a complete round in the track. The vanilla neural network was also able to make a complete round but it took longer time to learn. It was observed that the average reward per episode is around 40 for modified DQN and around 23 for vanilla neural network.

\begin{table}[!h]
\begin{center}
\begin{tabular}{ |c|c|c|c| } 
\hline
Model & Average reward & Training time (hrs) \\
\hline
Original DQN & 25 & 10 \\ 
Vanilla NN & 23 & 6 \\ 
Modified DQN & 40 & 4 \\ 
\hline
\end{tabular}
\caption{\label{model-comp} Performance of different models in terms of average reward and training time.}
\end{center}
\end{table}

\section{Conclusion}
In conclusion, we investigated the use of Deep Q-Networks (DQNs) in the context of a 2D self-driving car simulation. The simulation was designed to test the effectiveness of DQNs in learning an optimal policy for steering the car and avoiding obstacles.
The results of the simulation showed that DQNs can effectively learn to steer the car and avoid collisions in the simulated environment. However, the performance of the DQN was highly dependent on the action selection mechanism.

\section{Future Works}
In this project, we demonstrated the effectiveness of DQN for autonomous driving in 2D environment. However, there are several areas for future research that can build our findings and further improve the performance of our agent.
We can tune the hyperparameters of DQN model like learning rate, target network update frequency, replay buffer size, neural network architecture etc. Besides, we can use SUMO for simulating and analyzing vehicles in the map of University of Memphis which we couldn't do currently due to time constraints. It would be interesting to explore how our approach can be extended to handle multiple cars.

\balance{}

\bibliographystyle{SIGCHI-Reference-Format}
\bibliography{sample}

\end{document}